\documentclass[a4paper,11pt]{article}

\usepackage{fourier}
\usepackage{color}
 \usepackage{graphicx}
\usepackage{url}
\usepackage[affil-it]{authblk}
\usepackage{amsmath}

\usepackage[T1]{fontenc}
\usepackage{times}

\usepackage{algorithm}
\usepackage{algpseudocode}
\usepackage{latexsym}
\usepackage{cite}
\usepackage{enumitem}
\usepackage{amssymb,mathrsfs}
\usepackage{booktabs}
\usepackage{microtype}
\usepackage{siunitx}
\usepackage{dcolumn}
\usepackage{colortbl}
\usepackage{subfigure}
\usepackage{float}
\usepackage{color}
\usepackage[dvipsnames]{xcolor}
\usepackage{amsmath}
\usepackage{booktabs}
\usepackage{multirow}
\usepackage{wrapfig}
\usepackage{mathtools}

\DeclareMathOperator*{\jrargmax}{arg\,max}

\begin{document}

\title{Do the Frankenstein, or how to achieve better out-of-distribution performance with manifold mixing model soups}

\author{Hannes Fassold}
\affil{JOANNEUM RESEARCH - DIGITAL, Austria}
\date{}
\maketitle
\thispagestyle{empty}

\begin{abstract}

The standard recipe applied in transfer learning is to finetune a pretrained model on the task-specific dataset with different hyperparameter settings and pick the model with the highest accuracy on the validation dataset. Unfortunately, this leads to models which do not perform well under distribution shifts, e.g. when the model is given graphical sketches of the object as input instead of photos. In order to address this, we propose the \emph{manifold mixing model soup}, an algorithm which mixes together the latent space manifolds of multiple finetuned models in an optimal way in order to generate a fused model. We show that the fused model gives significantly better out-of-distribution performance (+3.5 \% compared to best individual model) when finetuning a CLIP model for image classification. In addition, it provides also better accuracy on the original dataset where the finetuning has been done. 
\end{abstract}


\section{Introduction}
\label{sec:manifoldmixms:intro}

Large pretrained visual foundation models like CLIP \cite{Radford2021-manifoldmixms} or CoCa \cite{Yu2022-manifoldmixms} got very popular recently due to their great performance for a variety of computer vision tasks, either as zero-shot learner (without finetuning) or serving as a base for task-specific finetuning on a smaller dataset. 

Typically, multiple models are finetuned with different hyperparameters (like learning rate, weight decay or data augmentation strategy), using the same pretrained model as initialization. From those, the model with the best accuracy on the validation dataset is selected. Unfortunately, this procedure leaves out important information which has been learned in the latent space manifolds (individual layers or a collection of layers) of the remaining finetuned models. As shown in \cite{Wortsman2022-manifoldmixms}, even fusing multiple finetuned models in a very straightforward way by averaging them makes the fused model already significantly more robust to distribution shifts in the data.

Motivated by this, we propose the \emph{manifold mixing model soup} (\emph{ManifoldMixMS}) algorithm. Instead of simple averaging, it uses a more sophisticated strategy to generate the fused model. Specifically, it partitions a neural network model into several latent space manifolds (which can be individual layers or a collection of layers). Afterwards, from the pool of finetuned models available after hyperparameter tuning, the most promising ones are selected and their latent space manifolds are mixed together individually. The optimal mixing coefficient for each latent space manifold is calculated automatically via invoking an optimization algorithm. The fused model we retrieve with this procedure can be thought as sort of a "Frankenstein" model, as it integrates (parts of) individual model components from multiple finetuned models into one model.

The remainder of the work is organized as follows. In section \ref{sec:manifoldmixms:stateofart} we revise related work. Section \ref{sec:manifoldmixms:algorithm} presents the proposed manifold mixing model soup algorithm. Section \ref{sec:manifoldmixms:evaluation} presents the experiments and evaluation, which show the advantage of the proposed algorithm with respect to the state of the art, especially with respect to distribution shifts in the data. Finally, section \ref{sec:manifoldmixms:conclusion} concludes the paper.

\section{State of the Art}
\label{sec:manifoldmixms:stateofart}

A variety of methods has been proposed recently for merging several models into one fused model, with the aim of increasing the generalization capability and robustness to distribution shifts of the fused model.

A classical method is \emph{stochastic weight averaging} (SWA) \cite{Izmailov2018-manifoldmixms}, which produces the fused model by averaging the weights of a set of models sampled from the final stages of a single training run. They show that SWA leads to solutions of the optimization problem that are wider than the optima found by standard SGD, which in turn leads to a better generalization of the fused model.

The authors of \cite{Oswald2021-manifoldmixms} propose to replicate and learn in parallel a subset of weights (e.g. the batch-norm and classification layers) in a late phase of neural network learning. These late-phase weights define an ensemble of models which share every other weight. These parameters are then optimized independently and subsequently averaged.

In \cite{Jolicoeur2023-manifoldmixms} an algorithm called \emph{population parameter averaging} (PAPA) is presented, which trains a population of models in parallel (with different learning rate, data augmentation strategies etc.). It improves overall performance by infrequently replacing the weights with the population average and frequently pushing all model weights slightly towards the population average. A disadvantage of this method is the high memory consumption, as the gradients for \emph{several} parallel training runs have to be kept up.

The work of \cite{Wortsman2022-manifoldmixms} shows that averaging the weights of multiple models finetuned with different hyperparameter configurations often improves accuracy and out-of-distribution performance of the averaged model. They propose two different averaging algorithms (which they call "souping"), \emph{uniform soup} and \emph{greedy soup}. The uniform soup is a very simple procedure, as it does an averaging of all finetuned models. In contrast, the greedy soup is constructed by sequentially adding each model as a potential ingredient to the soup, and only keeping the model if it improves the performance of the averaged model. Our proposed \emph{manifold maxing model soup} algorithm (see section \ref{sec:manifoldmixms:algorithm}) is inspired by their greedy soup algorithm. But in contrast to them, we are partitioning the model into several components (latent space manifolds) and do an optimization in order to calculate the optimal mixing factor for each component.

In \cite{Matena2022-manifoldmixms} it is shown that uniform averaging of several finetuned models corresponds to making an isotropic Gaussian approximation to their posteriors. The authors propose an alternative merging procedure based on the Laplace approximation, where each model's posterior is approximated as a Gaussian distribution whose precision matrix (inverse of the covariance matrix) corresponds to its Fisher information.

The authors of \cite{Wortsman2022a-manifoldmixms} found that while fine-tuning a pretrained vision model improves performance on the downstream task, it also tends to decrease accuracy on the original pretraining task. They therefore propose a robust finetuning procedure called \emph{WiSE-FT} that computes a weighted average of the original pretrained parameters and the finetuned parameters. Different weighting values produce different trade-offs between pretraining and finetuning task performance.


\section{Manifold mixing model soup}
\label{sec:manifoldmixms:algorithm}

In the following, we outline the proposed algorithm for generating a fused model -- the  \emph{manifold mixing model soup} -- from its ingredients (the finetuned models after hyperparameter tuning). The algorithm pseudocode can be seen in Algorithm \ref{alg:manifoldmixms:algorithm}.

We first sort all $n$ finetuned models $\theta_i$ (with $i=0,...,n-1$) in descending order, based on their validation accuracy $ValAcc(\theta_i)$ on the original dataset which was used for finetuning. So $\theta_0$ is the model (to be precise, its finetuned parameters) with the highest validation accuracy, whereas $\theta_{n-1}$ is the one with the lowest validation accuracy.

Each model $\theta_i$ is partitioned into $m$ components $\theta_i^j$, where $\theta_i^j$ corresponds to a single latent space manifold, and $j = 1,...,m$. Each latent space manifold comprises either a single layer or a collection of layers, corresponding to one building block of the model. Typically, we partition a model into $10-30$ components. A finer partitioning makes the subsequent optimization more difficult, whereas a too coarse partitioning provides not enough freedom to optimize the mixing of the latent space manifolds individually. The motivation for aggregating a collection of layers into one component is to reduce the number of variables during optimization, which makes it easier for the optimizer to find a good optimum. For each model, of course the same partitioning is employed. 

The fused model $\Psi$ is now calculated in an sequential way, by iteratively mixing promising ingredient models with it. At first, the fused model is set to the best finetuned model via $\Psi = \theta_0$, and the variable $k$, which counts the number of models which have been mixed so far into the fused model, is set to $1$.

In each iteration (for $i = 1,...,n-1)$, we try now to mix the candidate model $\theta_i$ with the current fused model $\Psi$ in an optimal way, with the aim of increasing the validation accuracy of the updated fused model $\Psi'$ (which includes $\theta_i$) on the original dataset. 

In order to save computation time, we skip the optimization step for a candidate model $\theta_i$ for which it is unlikely that we get an increase in the validation accuracy by mixing $\theta_i$ into the current fused model $\Psi$. For that, we generate the "approximate average" model $\tilde{\Psi}$ via
\begin{equation}
\label{eq:manifoldmixms:approxavg}
\tilde{\Psi} =  \frac{k}{k+1} \cdot \Psi + \frac{1}{k+1} \cdot \theta_i
\end{equation} 
and test whether the condition $ValAcc(\tilde{\Psi}) > \tau \cdot ValAcc(\Psi)$ is fulfilled. If so, we continue with this iteration. If it is not fulfilled, we skip the following steps of this iteration, so candidate model $\theta_i$ will not be taken into account. The motivation for the specific combination provided in Equation \eqref{eq:manifoldmixms:approxavg} is that $\tilde{\Psi}$ calculated in this way corresponds approximately to the \emph{average} of all candidate models (like in \cite{Wortsman2022-manifoldmixms}) which have been mixed so far into the fused model (including $\theta_i$), \emph{if we assume} that the optimization did not change the mixing coefficients drastically from their provided initial values. We set the constant $\tau$ to $0.998$.

Having identified $\theta_i$ as a promising candidate model, in the next step we determine the optimal factors for mixing its latent space manifolds into the current fused model $\Psi$. For this, we define the updated fused model $\Psi'(\lambda)$ as a \emph{component-wise} convex combination of $\Psi$ and $\theta_i$ via
\begin{equation}
\label{eq:manifoldmixms:convexcombfusedmodel}
\Psi'(\lambda)^j = \lambda^j \cdot \Psi^j + (1 - \lambda^j) \cdot \theta^j_i
\end{equation}
for all components $j = 1,...,m$. Note that $\Psi'(\lambda)$ is a function of the mixing  vector $\lambda$. The mixing factor $\lambda^j \in [0, 1]$ determines how much of the $j-th$ component (latent space manifold) of the candidate model $\theta_i$ is mixed into the current fused model $\Psi$. The component-wise convex combination of the two models allows an optimizer to explore the latent space manifolds of the models $\Psi$ and $\theta_i$ in a very flexible way, in order to find the optimal mixing vector $\lambda^* \in \mathbb{R}^m$ which gives the highest validation accuracy for the updated fused model $\Psi'$.

For the subsequent optimization step, we set up the optimization problem to solve as
\begin{equation}
\label{eq:manifoldmixms:optimprobem}
\lambda^* = \jrargmax_{\lambda \in [0,1]^m} \left( ValAcc \left( \Psi' \left( \lambda \right) \right) \right)
\end{equation}
where $[0,1]^m$ is the $m-$dimensional unit interval. Via the constraint $\lambda \in [0,1]^m$ we ensure that a convex combination is done for each component, so we are in fact \emph{interpolating linearly} between the latent space manifolds $\Psi^j$ and $\theta^j$. The model $\Psi'(\lambda)$ can be calculated via Equation \eqref{eq:manifoldmixms:convexcombfusedmodel}.

For solving this optimization problem, we employ the \emph{Nevergrad} \footnote{\url{https://facebookresearch.github.io/nevergrad/}}  optimization package. It provides a large variety of black-box \emph{derivative-free} optimization algorithms together with a sophisticated heuristic \cite{Liu2020-manifoldmixms} to select the best optimizer based on the characteristic (number of variables, allowed budget for function evaluations etc.) of the optimization problem. As the initial value for the mixing factors, we set $\lambda^j = k/(k+1)$ for $j = 1,...,m$ with a similar motivation as explained earlier for Equation \eqref{eq:manifoldmixms:approxavg}.

We invoke now the optimizer in order to calculate the optimal mixing vector $\lambda^*$ which give the highest validation accuracy on the dataset used for finetuning. The optimal updated fused model can be calculated now via ${\Psi'}^* = \Psi'(\lambda^*)$. We check now whether the condition $ValAcc({\Psi'}^*) > ValAcc(\Psi)$ is fulfilled. If so, we have found a better fused model ${\Psi'}^*$ by mixing $\theta_i$ into it. Consequently, we replace $\Psi$ by ${\Psi'}^*$ and increase $k$ by 1. If this is not the case, we keep the current fused model $\Psi$ and $k$ as they are.

After iterating over all candidate models $\theta_i$ for $i = 1,...,n-1$  we retrieve a final fused model $\Psi$ (the \emph{manifold mixing model soup}), which mixes together the $k$ selected candidate models / ingredients in an optimal way.

\begin{algorithm}[t]
\caption{Manifold mixing model soup algorithm}
\label{alg:manifoldmixms:algorithm}

\begin{algorithmic}

\renewcommand{\algorithmiccomment}[1]{{\normalsize\hfill \textcolor{CadetBlue}{ $\triangleright$ \emph{#1}}}}

\Require Finetuned models $\{\theta_0, ..., \theta_{n-1}\}$ as result of hyperparameter tuning
\Require Partitioning of a model $\zeta$ into $m$ components (latent space manifolds) $\zeta^j$ for $j = 1,...,m$
\Require Function $ValAcc(\zeta)$ which calculates validation accuracy for $\zeta$ on dataset used for finetuning 

\State $\{\theta_0, ..., \theta_{n-1}\} \gets sort(\{\theta_0, ..., \theta_{n-1}\})$ \Comment{ Sort $\{\theta_0, ..., \theta_{n-1}\}$ in descending order based on $Valacc(\theta_i)$ }
\State $k \gets 1$ \Comment{ Number of candidate models mixed into fused model}
\State $\Psi \gets \theta_0$ \Comment{ Set initial fused model to best finetuned model $\theta_0$ } 
\State $\tau \gets 0.998$ \Comment{ Tolerance factor for promising candidate models }

\For{$i = 1,...,n-1$} \Comment{ Iterative over all candidate models $\theta_i$}
  \State $\tilde{\Psi} =  \frac{k}{k+1} \cdot \Psi + \frac{1}{k+1} \cdot \theta_i$ \Comment{ Generate "approximate average" model }
  \If{$ValAcc(\tilde{\Psi}) > \tau \cdot ValAcc(\Psi)$} \Comment { Do optimization only if candidate is promising }
    \State $\Psi'(\lambda)^j = \lambda^j \cdot \Psi^j + (1 - \lambda^j) \cdot \theta^j_i$ \Comment { Define updated fused model for components $j = 1,...,m$ }
    \State $\lambda^* = \jrargmax_{\lambda \in [0,1]^m} \left( ValAcc \left( \Psi' \left( \lambda \right) \right) \right)$ \Comment{ Calculate optimal mixing factors}
    \State ${\Psi'}^* = \Psi'(\lambda^*)$ \Comment{ Calculate optimal updated fused model }
    \If{$ValAcc({\Psi'}^*) > ValAcc(\Psi$)}
      \State $k \gets k + 1$
      \State $\Psi \gets {\Psi'}^*$ \Comment{ Mix candidate model $\theta_i$ into current fused model }
    \EndIf
  \EndIf
\EndFor
\State \Return $\Psi$ \Comment{ Return final fused model }

\renewcommand{\algorithmiccomment}[1]
{{\normalsize\hfill$\triangleright$ #1}}

\end{algorithmic}

\end{algorithm}

\section{Experiments and Evaluation}
\label{sec:manifoldmixms:evaluation}


The setup for our experiments is very similar to the one for the vision models given in the \emph{model soup} paper \cite{Wortsman2022-manifoldmixms}. We summarize it in the following for clarity and completeness.

The model employed for finetuning is the \emph{CLIP} model \cite{Radford2021-manifoldmixms}.  CLIP is a powerful multi-modal zero-shot neural network, which has been pretrained with contrastive learning on a huge dataset of image-text pairs. Specifically, we use the \emph{CLIP ViT-B/32} variant specified in Table 20 of \cite{Radford2021-manifoldmixms} and provided in the \emph{OpenCLIP} package \footnote{\url{https://github.com/mlfoundations/open_clip}}. Finetuning of the pretrained model is performed end-to-end (all parameters are modified), as it typically leads to better performance than training only the final linear layer. Before finetuning, the final layer is initialized with a linear probe as described in \cite{Kumar2022-manifoldmixms}. The loss function employed for finetuning is the cross-entry loss.

The original dataset employed for finetuning is \emph{ImageNet} \cite{ImageNet-manifoldmixms}. Since the official ImageNet validation dataset is typically used as the test dataset, we use roughly $2 \%$ of the ImageNet training dataset as held-out validation dataset for calculating the validation accuracy in our proposed algorithm (see section \ref{sec:manifoldmixms:algorithm} and the pseudocode provided in Algorithm \ref{alg:manifoldmixms:algorithm}).

For measuring the out-of-distribution performance (robustness to distribution shifts) of our proposed algorithm, we employ five datasets derived from ImageNet with natural (not synthetically generated) distribution shifts. They corresponds to datasets with naturally occurring variations of the data samples due to different lighting, viewpoint, geographic location, image style (e.g. sketch instead of photo), crowdsourcing and more. The five datasets with distribution shifts we use are:
\begin{itemize}
 \item
ImageNet-V2 (IN-V2) \cite{ImageNetV2-manifoldmixms} is a reproduction of the ImageNet test set with distribution shift. The dataset was collected by closely following the original labelling protocol.
\item ImageNet-R (IN-R) \cite{ImageNetR-manifoldmixms} contains renditions (e.g., sculptures, paintings) for 200 ImageNet classes.
\item ImageNet-Sketch (IN-Sketch) \cite{ImageNetSketch-manifoldmixms} contains sketches instead of natural images. It contains only sketches in "black-and-white" color scheme.
\item ObjectNet \cite{ObjectNet-manifoldmixms} provides objects in various scenes with 113 classes overlapping with ImageNet.
\item ImageNet-A (IN-A) \cite{ImageNetA-manifoldmixms} is a test set of natural images misclassified by a ResNet-50 model for 200 ImageNet classes.
\end{itemize}

\begin{figure}[t]
    \centering
    \includegraphics[width = 1.0\textwidth]{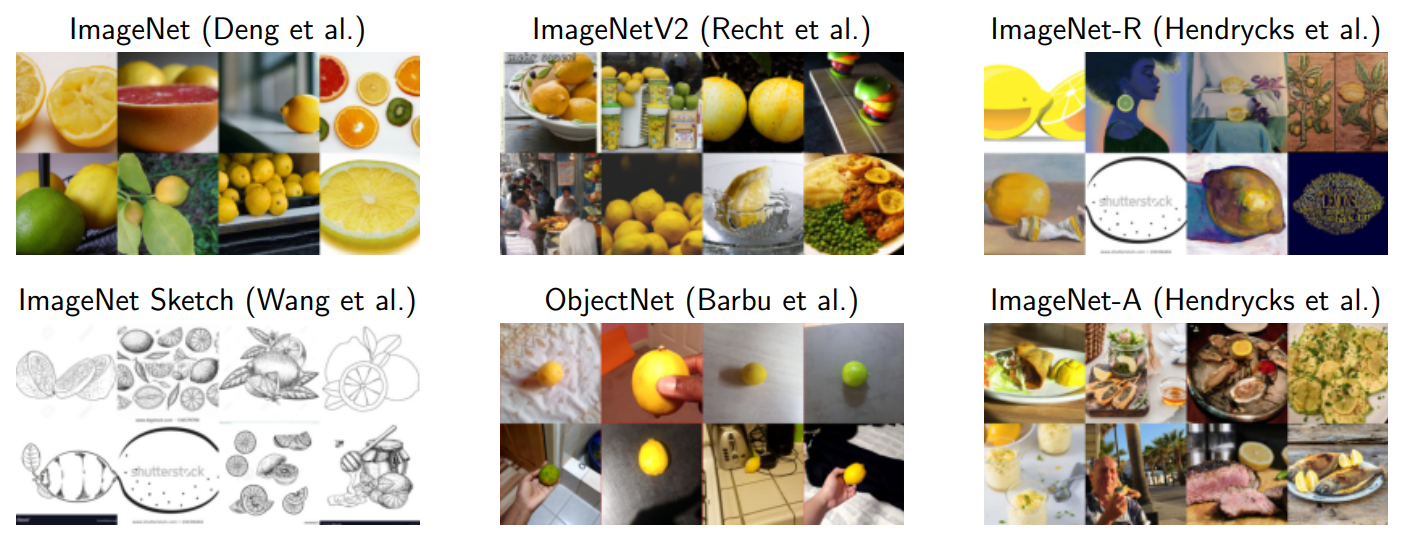}
    \caption{Samples for class \emph{lemon}, from the original ImageNet dataset and the five datasets with natural distribution shifts. Image courtesy of \cite{Wortsman2022a-manifoldmixms}}.
    \label{fig:manifoldmixms:ood_dataset_samples}
    \vspace{4mm}

\end{figure}

See Figure \ref{fig:manifoldmixms:ood_dataset_samples} for an illustration of samples for each of the datasets with natural distribution shifts. For all datasets (the original used for finetuning and the ones with distribution shifts), we take the top-1 accuracy on the respective test set for measuring the performance of a model. We calculate the overall out-of-distribution performance of a model as the average of its test accuracy over all five datasets with distribution shifts.

We partition the CLIP ViT-B/32 model into 8, 15 and 26 components. A too fine partitioning (e.g. one component for each layer of the model) makes the optimization much more difficult, whereas a too coarse partioning provides not enough flexibility for mixing the latent space manifolds individually in an optimal way. The structure of the partitioning is done roughly according to the hierarchy of the building blocks of the CLIP model. We denote the respective variant of our proposed algorithm with 8, 15 and 26 components as ManifoldMixMS-C8, ManifoldMixMS-C15 and ManifoldMixMS-C26. 

We parametrize the Nevergrad optimizer with a maximum budget for the number of function evaluations (of the objective function to optimize) of roughly $250$ function evaluations for all ManifoldMixMS variants. The employed optimizer is automatically selected by the Nevergrad optimization package (see \cite{Liu2020-manifoldmixms}). For our cases, it always selects the \emph{Cobyla} \cite{Powell1994-manifoldmixms} optimization algorithm. The Cobyla algorithm is one of the best derivative-free algorithms for optimization of continuous variables with bound constraints, especially when the allowed number of function evaluations is quite small.

For the evaluation of our proposed manifold mixing model soup algorithm, we compare mainly with the \emph{greedy soup} and \emph{uniform soup} algorithms which have been proposed in \cite{Wortsman2022-manifoldmixms}. Additionally, we compare our proposed algorithm also against ensemble models. We compare against the same ensemble models as done in \cite{Wortsman2022-manifoldmixms} and take also the accuracy numbers reported there for them. Of course, one should take into account that the computational cost for inference of an ensemble model is much higher -- $K$ times higher for an ensemble model consisting of $K$ individual models -- than for our proposed ManifoldMixMS algorithm which produces only a single fused model.

\begin{figure}[t]
    \centering
    \includegraphics[width = 0.85\textwidth]{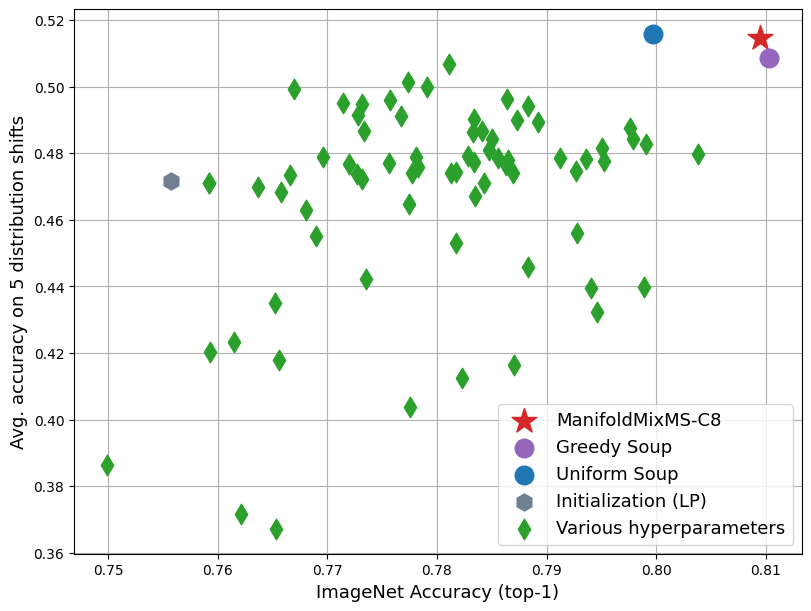}
    \caption{Comparison of our proposed manifold mixing soup algorithm (with 8 components) against greedy soup and uniform soup algorithm from \cite{Wortsman2022-manifoldmixms} and the individual finetuned models.}
    \label{fig:manifoldmixms:scatterplot}
    
\end{figure}

The scatterplot in Figure \ref{fig:manifoldmixms:scatterplot} shows how our proposed ManifoldMixMS-C8 algorithm (the overall best variant) performs compared to the greedy soup and uniform soup algorithm from \cite{Wortsman2022-manifoldmixms} and to the individual finetuned models. 

Furthermore, Table \ref{tab:manifoldmixs:eval} gives a detailed evaluation of our proposed variants of the manifold mixing soup algorithm with 8, 15 and 26 on the five datasets with distribution shifts (ImageNet-V2, ImageNet-R, ImageNet-Sketch, ObjectNet, ImageNet-A) as well as on the original dataset used for finetuning (ImageNet).

One can see clearly from the scatterplot that our proposed manifold mixing model soup (especially the preferred variant with 8 components) algorithm combines the best properties of the uniform model soup and greedy soup algorithm. Specifically, it has practically the same good out-of-distribution accuracy as the uniform soup algorithm and still keeps the good accuracy of the greedy soup algorithm on the original ImageNet dataset. In contrast, the uniform soup algorithm performs on the original ImageNet dataset even worse than the best individual finetuned model.

It is significantly better with respect to the best finetuned model both on the datasets with distribution shifts ($+3.5\%$), but also on the original ImageNet dataset ($+0.6\%$). The difference grows even bigger when comparing with the second-best finetuned model.

Surprisingly, it has also a significantly better out-of-distribution accuracy than both Ensemble methods, although its accuracy on the original ImageNet dataset is worse especially when compared with the greedy ensemble method. As already mentioned, one should take into account that Ensemble methods have a much higher computational cost.


\newcommand\JrsManBlue{\color{NavyBlue}} 
\newcommand\JrsManRed{\color{Maroon}}

\begin{table*}
            \caption{Detailed comparison of our proposed manifold mixing soup algorithm variants for the CLIP ViT-B/32 neural network with the best and second-best finetuned model, the model soup algorithms from \cite{Wortsman2022-manifoldmixms} and for completeness also with Ensemble methods.The top-1 accuracy (in \%) on the respective test dataset is employed. The column "Avg OOD" corresponds to the average over all 5 datasets with distribution shifts. The best and second-best result for each dataset (without taking into account the Ensemble methods as they have a much higher computational cost) is marked in {\color{Maroon}red} and {\color{NavyBlue}blue}. 
            }
            \vspace*{4mm}
            \label{tab:manifoldmixs:eval}
            \centering
            \setlength{\tabcolsep}{0.4em}
            \begin{tabular}{ l | r | r r r r r | r}
                \toprule

                \multicolumn{1}{c|}{Method} & ImageNet  & \multicolumn{1}{c}{IN-V2} & \multicolumn{1}{c}{IN-R} & \multicolumn{1}{c}{IN-Sketch} & \multicolumn{1}{c}{ObjectNet} & \multicolumn{1}{c|}{IN-A} & \multicolumn{1}{c}{Avg OOD}\\

                \midrule

                Best finetuned model & 80.38 & 68.44 & 44.51 & 60.63 & 42.62 & 23.64 & 47.97 \\
                Second-best finetuned model & 79.89 & 67.91 & 41.49 & 54.58 & 37.98 & 18.01 & 44.01 \\
                \midrule
                Uniform soup & 79.97 & 68.51 & 47.71 & \JrsManRed{66.54} & \JrsManRed{45.95} & \JrsManRed{29.17} & \JrsManRed{51.57} \\
                Greedy soup & \JrsManRed{81.03} & 69.55 & 47.77 & 64.20 & 44.90 & 27.89 & 50.86 \\
                \midrule
                ManifoldMixMS-C8 & \JrsManBlue{80.95} & \JrsManRed{69.67} & \JrsManRed{48.15} & \JrsManBlue{64.81} & 45.66 & \JrsManBlue{29.06} & \JrsManBlue{51.47} \\         
                ManifoldMixMS-C15 & 80.80 & \JrsManBlue{69.61} & 47.89 & 64.76 & 44.45 & 28.39 & 51.02 \\                
                ManifoldMixMS-C26 & 80.85 & 69.58 & \JrsManBlue{48.04} & 64.79 & \JrsManBlue{45.75} & 28.88 & 51.41 \\
                \midrule
                Ensemble & 81.19 & -- & -- & -- & -- & -- & 50.77 \\
                Greedy ensemble & 81.90 & -- & -- & -- & -- & -- & 49.44 \\
                \bottomrule
            \end{tabular}
            \end{table*}

\section{Conclusion}
\label{sec:manifoldmixms:conclusion}

We propose the \emph{manifold mixing model soup} algorithm, which mixes together the latent space manifolds of multiple finetuned models in an optimal way in order to generate a fused model. Experiments show that the fused model gives
significantly better out-of-distribution performance (+3.5 \% compared to best finetuned model) when finetuning a CLIP model for image classification.

In the future, we plan to evaluate the proposed algorithm on other neural network architectures, for both computer vision as well as natural language processing tasks. Furthermore, we plan to do a theoretical analysis of the properties of the proposed algorithm in order to get a better insight why it provides a better out-of-distribution performance.

\section*{Acknowledgment}
\label{sec:manifoldmixms:acknowledgment}

This work was supported by European Union´s Horizon 2020 research and innovation programme under grant number 951911 - AI4Media.

\bibliographystyle{apalike}

\bibliography{imvip}

\end{document}